\documentclass[letterpaper, 10 pt, conference]{ieeeconf}  

\usepackage{graphicx}
\usepackage{subcaption}
\usepackage{mathtools}
\usepackage{bm}
\usepackage{esvect}
\usepackage{amsmath}
\usepackage{amssymb}
\usepackage{float}
\usepackage[font=footnotesize]{caption}
\usepackage{xcolor}

\IEEEoverridecommandlockouts                              

\title{\LARGE \bf
Path tracking control of self-reconfigurable robot hTetro with four differential drive units
}
\author{Yuyao Shi$^{1}$, 
        Mohan Rajesh Elara$^{2}$,
        Anh Vu Le$^{3}$,
        Veerajagadheswar Prabakaran$^{4}$
        and Kristin L. Wood$^{5}$
\thanks{*This work is supported by Singapore National Robotics R\&D Program Office and SUTD-MIT International Design Center.}
\thanks{$^{1}$Y. Shi is with the Department
of Engineering Product Development (EPD), Singapore University of Technology and Design (SUTD), Singapore, 487372,
        {\tt\small yuyao\_shi@sutd.edu.sg}}%
\thanks{$^{2}$M. R. Elara is with Department of EPD, SUTD, Singapore, 487372,
        {\tt\small rajesh\_mohan@sutd.edu.sg}}%
\thanks{$^{3}$A. V. Le is with Optoelectronics Research Group, Faculty of Electrical and Electronics Engineering, Ton Duc Thang University, Ho Chi Minh City 700000, Vietnam,
        {\tt\small leanhvu@tdtu.edu.vn}}%
\thanks{$^{4}$V. Prabakaran is with Department of EPD, SUTD, Singapore, 487372,
        {\tt\small prabakaran@sutd.edu.sg}}%
\thanks{$^{5}$K. L. Wood is the associate provost (graduate studies), Co-director of SUTD-MIT International Design Centre, SUTD, Singapore, 487372,
        {\tt\small kristinwood@sutd.edu.sg}}%
}

\begin{document}

\maketitle
\thispagestyle{empty}
\pagestyle{empty}

\begin{abstract}
The research interest in mobile robots with independent steering wheels has been increasing over recent years due to their high mobility and better payload capacity over the systems using omnidirectional wheels. However, with more controllable degrees of freedom, almost all of the platforms include redundancy, which are modeled using the instantaneous center of rotation (ICR). 
This paper deals with a Tetris-inspired floor cleaning robot hTetro which consists of four interconnected differential-drive units. Each module has a differential drive unit which can steer individually. 
Differing from most other steerable wheeled mobile robots, the wheel arrangement of this robot changes because of its self-reconfigurability. 
In this paper, we proposed a path tracking controller that can handle discontinuous trajectories and sudden orientation changes for hTetro. Singularity problems are resolved on both the mechanical aspect and the control aspect. The controller is tested experimentally with the self-reconfigurable robotic platform hTetro, and results are discussed.
\end{abstract}
\section{INTRODUCTION}\label{intro}

\begin{figure}[thpb]
\captionsetup[subfigure]{font=scriptsize}
\begin{subfigure}{0.25\textwidth}
\centering
\vspace*{0.08in}
\includegraphics[width=3.4in]{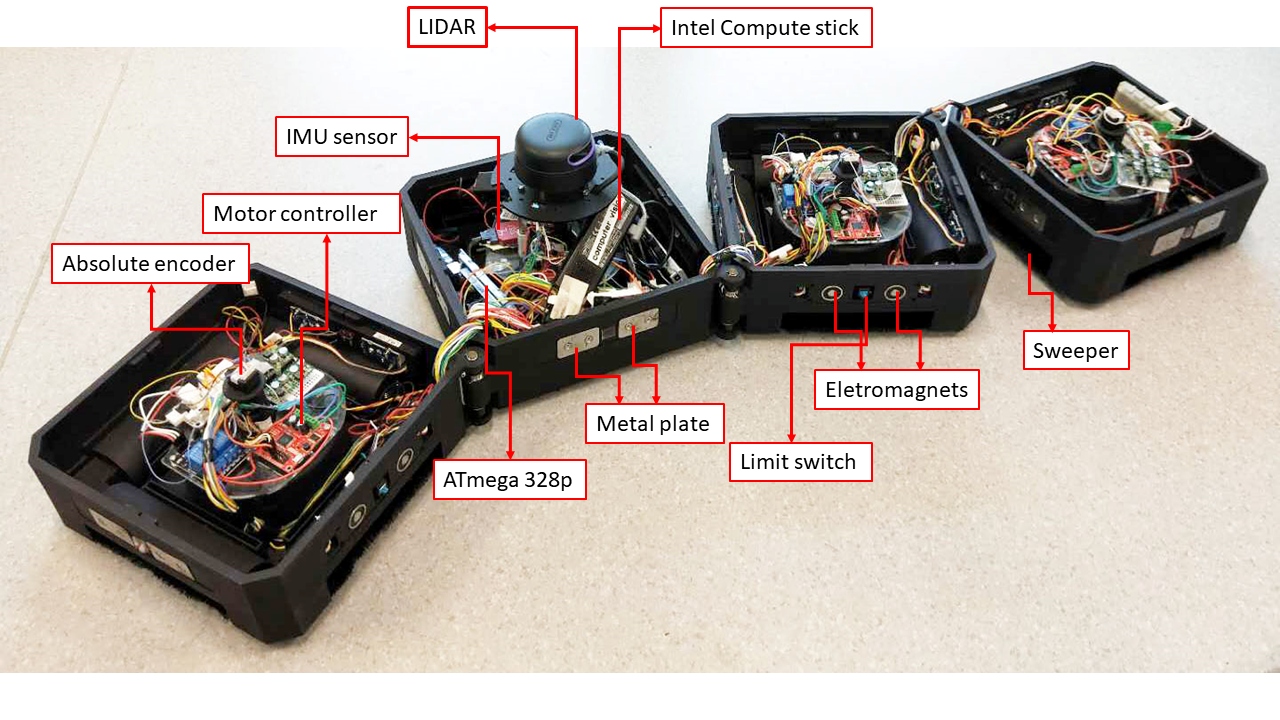}
\caption{Hardware design of the current hTetro}
\label{fig:subhardware}
\end{subfigure}
\begin{subfigure}[thpb]{0.25\textwidth}
\centering
\includegraphics[width=2in]{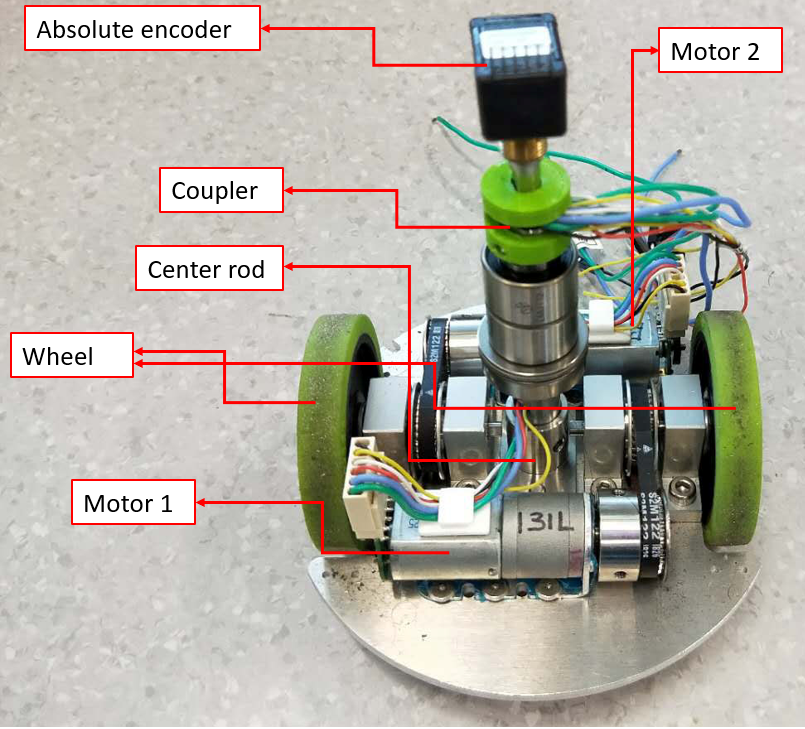}
\caption{Hardware design of the locomotion module}
\label{fig:sublocomotion}
\end{subfigure}%
~
\begin{subfigure}[thpb]{0.25\textwidth}
\centering
\includegraphics[width=1.2in]{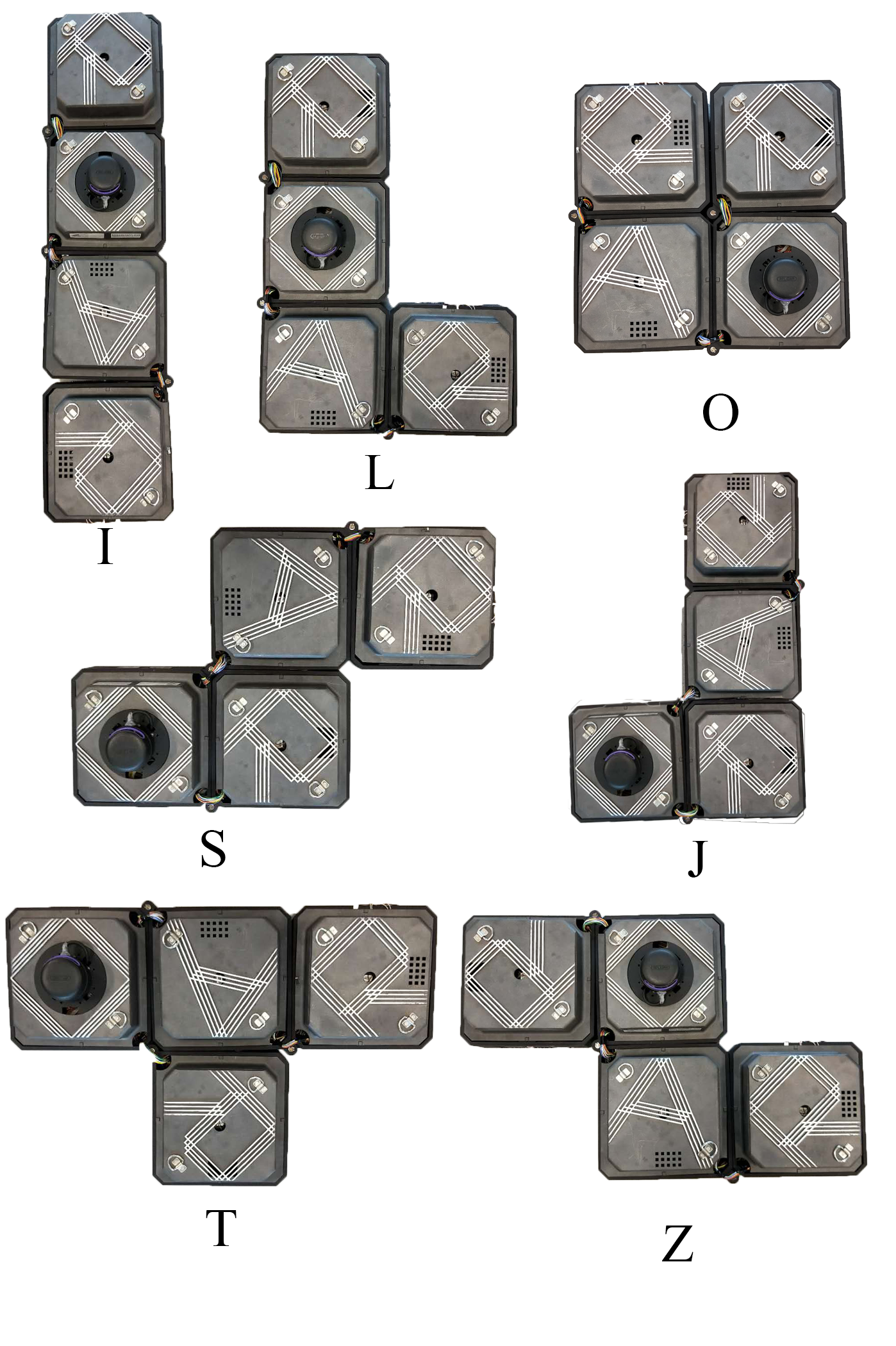}
\caption{7 configurations of hTetro}
\label{fig:subconfiguration}
\end{subfigure}
\caption{Mechanical design and hardware placement}
\label{fig_hardware}
\end{figure}

 The concept of modular and self-reconfigurable mechanisms are widely used due to its robustness, adaptability and multi-functionality. Self-reconfigurable robots are usually composed of multiple modules and able to perform shape-shifting according to different circumstances. A variety of reconfigurable systems have been developed to complete a wide range of tasks \cite{Fei2014}\cite{10.1115/1.3125205}. The system addressed in this paper is a self-reconfigurable robot developed for the floor-cleaning purpose. Inspired from the Tetris game, hTetro was designed to have four modular blocks, which allow hTetro to transform into seven configurations as shown in Fig. \ref{fig:subconfiguration}. This enables it to access narrow spaces and increases the area coverage as proved in \cite{8387830}\cite{7989725}. 
The advantages of using linked modular-type robot over multiple independent units are 1) the former is not limited by means of data transfer as the signals are transmitted through wires. 2) The requirement of computational power is lower for controlling one unit comparing to multiple units. Thus, with the same ability in area coverage, the self-reconfigurable robotic architecture has lesser requirements for the sensors, communication modules, actuators and other hardware. The platform introduced in this paper is a robust version of hTetro which uses four differential drive modules and free hinges instead of mecanum wheels with actuated servo hinges \cite{8387830}. By using a differential drive against four mecanum wheels in each module \cite{8387830}, hTetro is capable of moving on uneven terrain such as carpet or cemented granular. In \cite{8387830}\cite{7989725}, active hinge joints were used to change the configuration, which results in using of additional servos for reconfiguration and leads to low robustness due to servo break down. The modified design presented here uses free hinged joints with suspension. Electromagnets are used to maintain the specific configuration during locomotion while transformation is done by decoupling electromagnets and driving the modules individually. The independent steering action in each of the four modules gained using differential wheel action. \cite{7989725}.

The difficulty in controlling a robot with multiple steering modules lies upon satisfying the rigid body kinematic constraints. Lee and Tzuu-Hseng proposed a kinematic and dynamic control law based on the Lyapunov method and verified the stability in simulation \cite{8412839}. 
The drawback with this method is that it does not consider the non-holonomic constraints of the platform during the modeling which in practice will lead to improper alignment of wheels and cause tire wear because of the skidding motion. Some of the solutions are to design mechanical linkages between the wheels to prevent conflict in steering angles \cite{5153343}\cite{inproceedings1}\cite{8876854}. However, the additional mechanical linkage limits the steering angle and results in lesser maneuverability. It is also not suitable in this reconfigurable platform.

Modeling the platform under the instantaneous center of rotation (ICR) space is an alternative method. 

Although using the ICR and individual steering wheel configuration ensures the stability, there are singularity problems associated with it. The singularity happens on two aspects, namely, representation aspect and numerical aspect. The former refers to the situation where the steerable wheels are parallel with each other. In this case, the ICR is theoretically located at infinity. In order to solve the representational singularity, coordinate switching is one of the proposed methods \cite{5152450}\cite{6386131}. The latter refers to the situation where the ICR is located along the steering axes. In this case, there are more than one solutions to the desired angular velocity. This singularity has been addressed using artificial potential fields, as reported in \cite{5152450}\cite{5979549}\cite{5652685}\cite{Khatib1986}
or by offsetting the steering axis outside of the wheel plane \cite{7781653}\cite{7487360}\cite{SOROUR2019131}. However, the usage of artificial potential field limited the position of ICR while offsetting the steering axis leads to the coupling between steering and propulsion motion \cite{inproceedings}. In all the method proposed, the control problem is complicated either because of considering the coupling between steering and propulsion motion or involving dynamic control \cite{6907280}\cite{7989283}\cite{inproceedings1}. However, in this approach, the usage of differential drive solved the numerical singularity while not coupling the steering and propulsion motion. Therefore, the controller proposed in this paper is only based on kinematics of the robot which is simpler than dynamic control.  
 
The earlier version of hTetro also made use of the individual steering drive mechanism \cite{TUN2019102796}. Predefined-Radius-Locomotion (PRL) and Fixed-Heading-Angle-Locomotion (FHAL) were performed to test the maneuverability of the robot. The transformation mechanism of hTetro was also illustrated in detail. However, no path tracking control strategy and singularity avoidance were considered. 

The main contributions of this paper focus on two aspects. Firstly, a mechanical design that avoids the singularity where ICR is located at the steering center. By using differential drive, the numerical singularity problem can be avoided naturally without complicating the kinematics. Secondly, a path tracking controller that fits different wheel arrangements. The controller can be considered as two layers: a higher level controller to determine the ICR and the corresponding steering angles, and a lower level controller on the module level to ensure the kinematic constraints during transient state. 

The paper is organised into five sections. The mechanical design of hTetro and its kinematic model based on instantaneous center of rotation is presented in Section \ref{second}. Section \ref{control} describes the controller design and the method to avoid conflicting the kinematic constraints. The robust mechanical design and control is supported by experiments in Section \ref{experiment}. Section \ref{conclusion} concludes with future works.

\section {KINEMATIC MODEL}\label{second}

\begin{figure}[!t]
\centering
\includegraphics[width=3.3in]{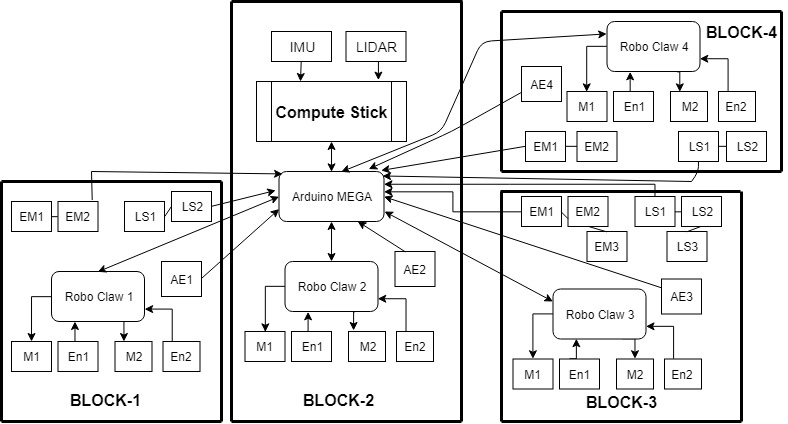}
\caption{hTetro System Architecture}
\label{fig:system Arch}
\end{figure}
In this section, the mechanical design and the system architecture of the current hTetro platform are presented. A kinematic formulation is also developed in Section \ref{kinematics} which is different from that of the normal fixed morphology robot because of its reconfigurability.

\subsection {Robot description and modeling parameters}\label{hardware}
\begin{figure}[!t]
\captionsetup[subfigure]{font=scriptsize}
\begin{subfigure}{0.4\textwidth}
\centering
\vspace*{0.1in}
\includegraphics[width=3in]{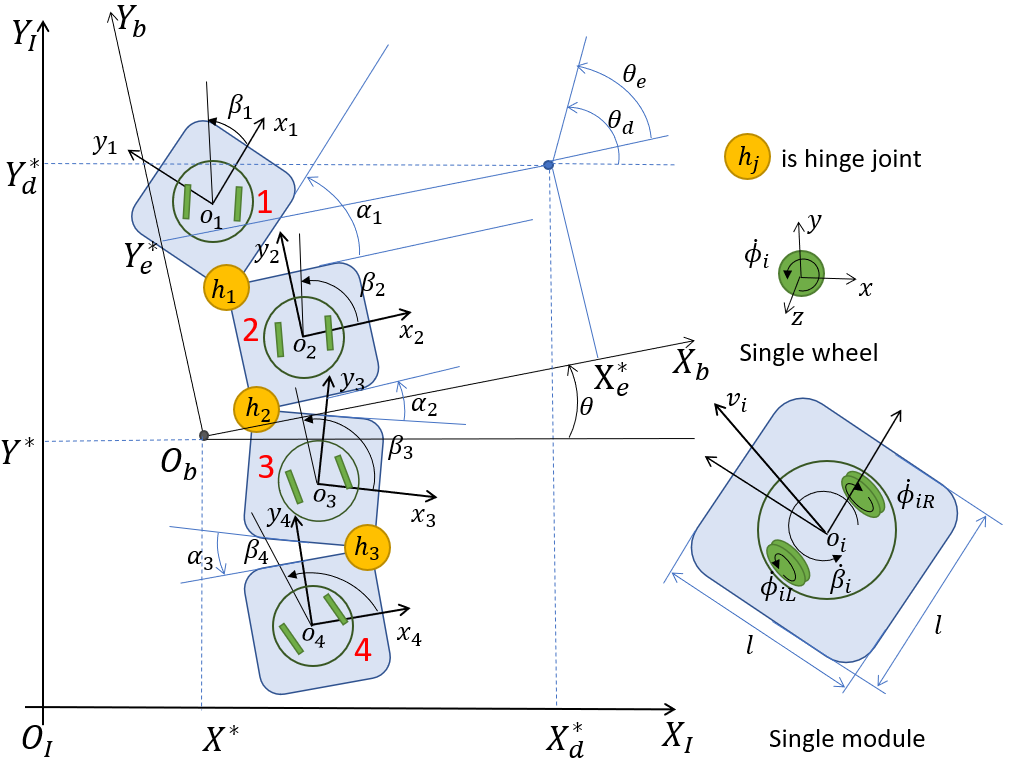} 
\caption{Schematic of the robot kinematic model}
\label{fig:subim1}
\end{subfigure}

\begin{subfigure}{0.4\textwidth}
\centering
\includegraphics[width=3in]{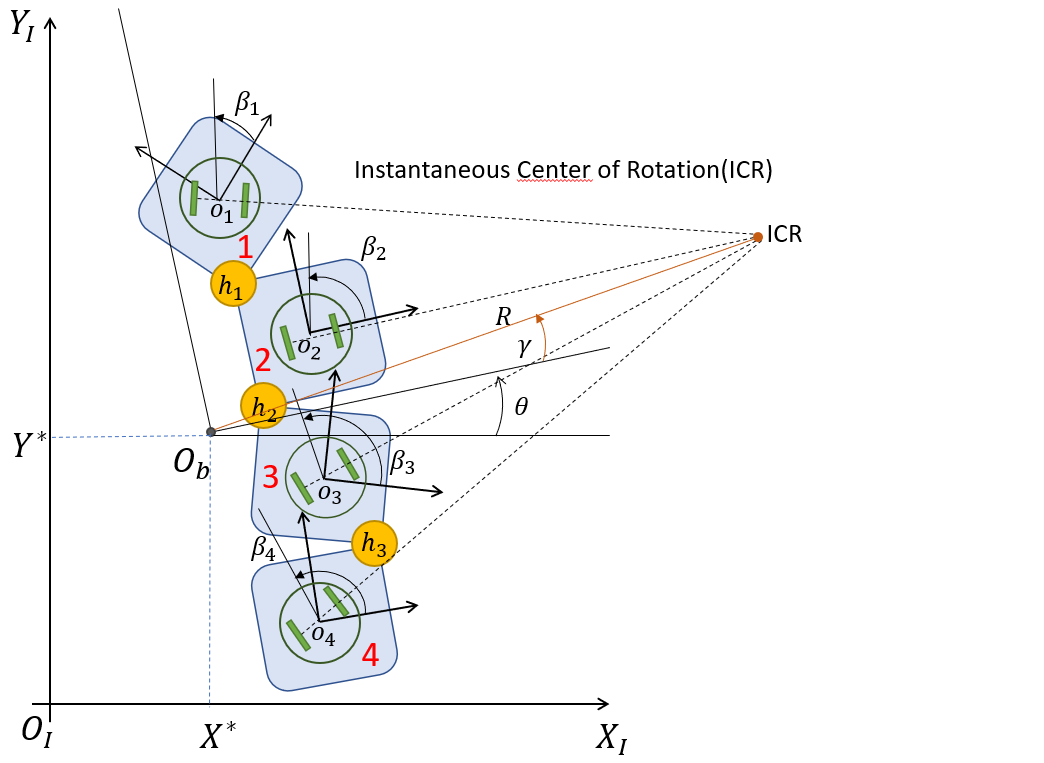} 
\caption{Schematic of ICR model}
\label{fig:subim2}
\end{subfigure}
\caption{Schematic diagrams of hTetro.}
\label{fig_schematic}
\end{figure}

The platform, as shown in Fig. \ref{fig:subhardware}, consists of four modules connected by three free hinges with suspensions. Each module has a differential-drive locomotion module driven by two 12V DC motors. The locomotion modules are connected with the four chassis through the bushing on the central rod. Hence, the heading angles of the locomotion modules are independent of the orientations of their corresponding chassis as well as that of the whole robot. With respect to electronics, we separated it as global and local peripheries. The global components such as LIDAR, IMU, and compute stick act as a top layer which helps the robot to perform simultaneous localization and mapping (SLAM), and autonomous navigation. Since block 2 acts as an anchor point for hTetro, we align the orientation of the whole robot with that of the second block. The local components such as motor driver (Roboclaw), electromagnet (EM), limit switch (LS), absolute encoders (AE) motors (M), and its encoders (En) would execute the reconfiguration and locomotion of hTetro. Electromagnets were used to maintain the morphology during navigation. Limit switches were used to ensure the completion of reconfiguration and to detect the morphology failures. Except in block 2, we housed two sets of electromagnets and limit switches in block 1, and block 4 and three sets in block 3. For locomotion, we have a set of motor driver, motor, and encoders in each block. We used Roboclaw motor driver to control the motors and to collect the encoder signals which would be transferred to the local controller. On top of every locomotion module, we housed an absolute encoder (AE) that could sense the heading angle of each locomotion module and send the data back to the controller. Almost every local peripheral's data acquisition is done by the local controller for which we employed Arduino MEGA that is housed in block 2. Arduino Mega is the only local component that communicates with compute stick and distributes commands to low-level peripheries to perform the tasks effectively. The detailed system architecture of hTetro is shown in Fig. \ref{fig:system Arch}. The assumptions made during the modeling are 1) The robot does not perform shape changing during locomotion. In other words, it is treated as a rigid platform during the locomotion, 2) The wheels roll with no slipping and no skidding during the locomotion. The legends for the symbols used are described in TABLE \ref{parameters} (see also Fig. \ref{fig_schematic}).
\begin{table}[!t]
\vspace*{0.1in}
\renewcommand{\arraystretch}{1.3}
\caption{modeling Parameters}
\label{parameters}
\centering
\begin{tabular}{c l}
\hline
Symbols & Description\\
\hline \hline
$\mathcal{F_I}$ & World frame\\
$\mathcal{F}_b$ & Robot frame\\
$X^*$, $Y^*$ &  Current position of COG w.r.t. frame $\mathcal{F}_I$\\
$X_d^*$, $Y_d^*$ & Desire position of COG w.r.t. frame $\mathcal{F}_I$\\
$\theta^*$ & Current orientation of the robot w.r.t. frame $\mathcal{F}_I$\\
$\theta_d^*$ & Desire orientation of the robot w.r.t. frame $\mathcal{F}_I$\\
$\theta_e$ & Orientation error\\
$\mathcal{F}_{bi}$ & Individual frame at steering axis of $i^{th}$ module\\
$\beta_i$ & Steering angle w.r.t. chasis-fixed frame $\mathcal{F}_{bi}$\\
$\alpha_j$ & Angle at $j^{th}$ hinge\\
$X_e^*, Y_e^*$ & Position error under $\mathcal{F}_b$\\
$R$ & Current instantaneous radius of rotation under $\mathcal{F}_b$\\
$\gamma$ & Current driving angle of the robot under $\mathcal{F}_b$\\
$R_d$ & Desire instantaneous radius of rotation under $\mathcal{F}_b$\\
$\gamma_d$ & Desire driving angle of the robot under $\mathcal{F}_b$\\
$\phi_{iL}$, $\phi_{iR}$ & rotational speed of $i^{th}$ module's left/right wheel\\
$v_i$ & Linear speed of $i^{th}$ module\\
$r_w$ & Wheel radius\\
$d$ & Wheel to center distance\\
$l$ & Length of each module\\
\hline 
\end{tabular}
\end{table}
\subsection {Kinematic modeling}\label{kinematics}
Fig. \ref{fig_schematic} shows a schematic plot of the mobile robot w.r.t. the world inertial frame $\mathcal{F_I}$. The position of the robot is defined as the position of the centroid from the top view and the orientation is defined by the orientation of LIDAR which imposes the orientation of the chassis of the second module. Each individual module is separated from the others. So for each differential drive module, it has $v_i =\frac{1}{2}r_w(\phi_{iL}-\phi_{iR})$ and $\dot{\beta}_i = \frac{1}{2}r_w(\phi_{iL}+\phi_{iR})$. Let $\boldsymbol{\xi}^* = [x^*\:y^*\:\theta^*]^T$ be the 3$\mathcal{D}$ task space coordinate under $\mathcal{F}_I$ and $\boldsymbol{\xi_d}^*$ denotes the desire one. The position of the centroid of the platform $O_b$ is defined to be $[\frac{\sum^4_{i=1}x_i}{4},\frac{\sum^4_{i=1}y_i}{4}]$ where $x_i$ and $y_i$ are the position of the steering axis of each block $i$ w.r.t. the inertial frame ($\mathcal{F_I}$). By treating each locomotion module as a whole, the forward kinematics of the robot can be expressed as:
\begin{equation} \label{foward_kinematics}
\dot{\boldsymbol{\xi}}^* = 
\mathbf{R}(\theta)\mathbf{G}\boldsymbol{v}
\end{equation}
with $\boldsymbol{v} = [v_1 \quad v_2 \quad v_3 \quad v_4]^T$ and $\mathbf{G}^T =$ 
\[\left [
\begin{tabular}{ccc}
$\cos(\beta_1+\alpha_1)$ & $\sin(\beta_1+\alpha_1)$ & $\frac{\vv{O_bO_1}^{\perp}}{4||\vv{O_bO_1}||^2}$\\
$\cos(\beta_2)$ & $\sin(\beta_2)$ & $\frac{\vv{O_bO_2}^{\perp}}{4||\vv{O_bO_2}||^2}$\\
$\cos(\beta_3+\alpha_2)$ & $\sin(\beta_3+\alpha_2)$ & $\frac{\vv{O_bO_3}^{\perp}}{4||\vv{O_bO_3}||^2}$\\
$\cos(\beta_4+\alpha_2+\alpha_3)$ & $\sin(\beta_4+\alpha_2+\alpha_3)$ & $\frac{\vv{_bO_4}^{\perp}}{4||\vv{O_bO_4}||^2}$\\
\end{tabular}
\right ] \]

The last column of $\mathbf{G}^T$ is to get the orthogonal projection of $\boldsymbol{v}$ on the line connecting the centroid of the platform and that of each module.
$\mathbf{R}(\theta)$ is the transformation matrices from $\mathcal{F}_b$ to $\mathcal{F_I}$.

 In addition, with the assumption that no reconfiguration is allowed during locomotion, $\dot{\theta}^*$ is defined to be the angular velocity at the centroid of the current shape which coincide with the angular velocities of all four chassis. The inverse kinematics is found through pseudo inverse of the forward kinematics.

However, as the platform is a redundant system, only using pseudo inverse to get the inverse kinematics does not ensure that the kinematic constraints are fulfilled. Therefore, the actuation command of each motor is regulated with respect to the ICR placement which will be discussed in the following section.

\section{PATH TRACKING CONTROLLER DESIGN}\label{control}
\begin{figure}[!t]
\centering
\vspace*{0.1in}
\includegraphics[width=3.3in]{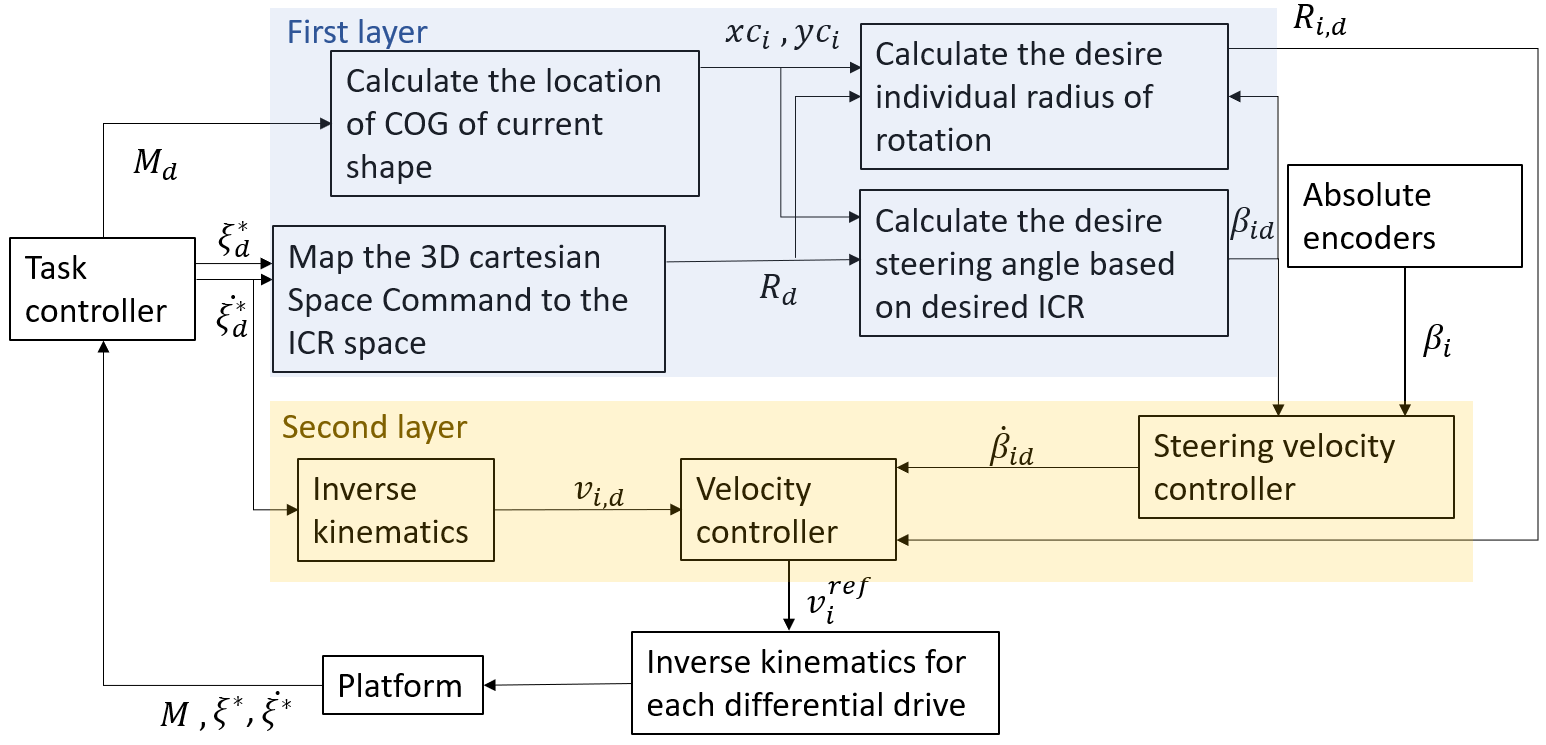}
\caption{Control framework of hTetro}
\label{fig_framework}
\end{figure}

In this section, the control framework for hTetro are presented, including locating of ICR, steering angles calculation, and controller design. The schematic diagram  of the control framework is shown in Fig. \ref{fig_framework}. In order to perform the cleaning task, the platform navigation is based on waypoints which contain the information of the desired position and orientation. Only when the platform reaches the waypoint correctly, the next waypoint information will be given. In between two waypoints, the centroid of the robot is required to follow a straight path while tracking the desired orientation. During the process above, the desired position and velocity ($\boldsymbol{\xi}^*$, $\dot{\boldsymbol{\xi}}^*$) were given by a high-level perception controller and then mapped to find the desired location of instantaneous center of rotation ($ICR_d$). Inspired from the previous work \cite{6386131}, the position of the $ICR_d$ is expressed using polar coordinates assigned with $\mathcal{F}_b$. The output is then used to calculate the desired corresponding steering angles ($\beta_{id}$), and individual radius of rotation ($r_i$). These information are fed into the individual steering angle controller together with the current instantaneous center of rotation to determine the angular velocity ($\dot{\beta}_{id}$) of each steering module. After that, due to the speed limit of the motor, a velocity controller will be used to regulate the desired linear velocity of each module ($v_{i,d}$) to generate the corresponding angular velocity ($\dot{\phi}_{iL}^{real}$ and $\dot{\phi}_{iR}^{real}$) of each wheel. 

\subsection{First Layer: Locating Desired Instantaneous Center of Rotation}\label{ICR}
As shown in Fig. \ref{fig_framework}, desired position $(X_d^*, Y_d^*)$ on the reference trajectory and current position $(X^*, Y^*)$, the error defined under robot frame is defined by: 
\begin{equation}\label{error}
\left [
\begin{tabular}{c}
$X_e^*$ \\
$Y_e^*$
\end{tabular}
\right ]
= 
\left [
\begin{tabular}{cc}
$\cos(\theta^*)$ & $-\sin(\theta^*)$\\
$\sin(\theta^*)$ & $\cos(\theta^*)$
\end{tabular}
\right ]\left [
\begin{tabular}{c}
$X_d^* - X^*$\\
$Y_d^* - Y^*$
\end{tabular}
\right ]
\end{equation}

\begin{figure}[!t]
\centering
\vspace*{0.1in}
\includegraphics[width=3.5in]{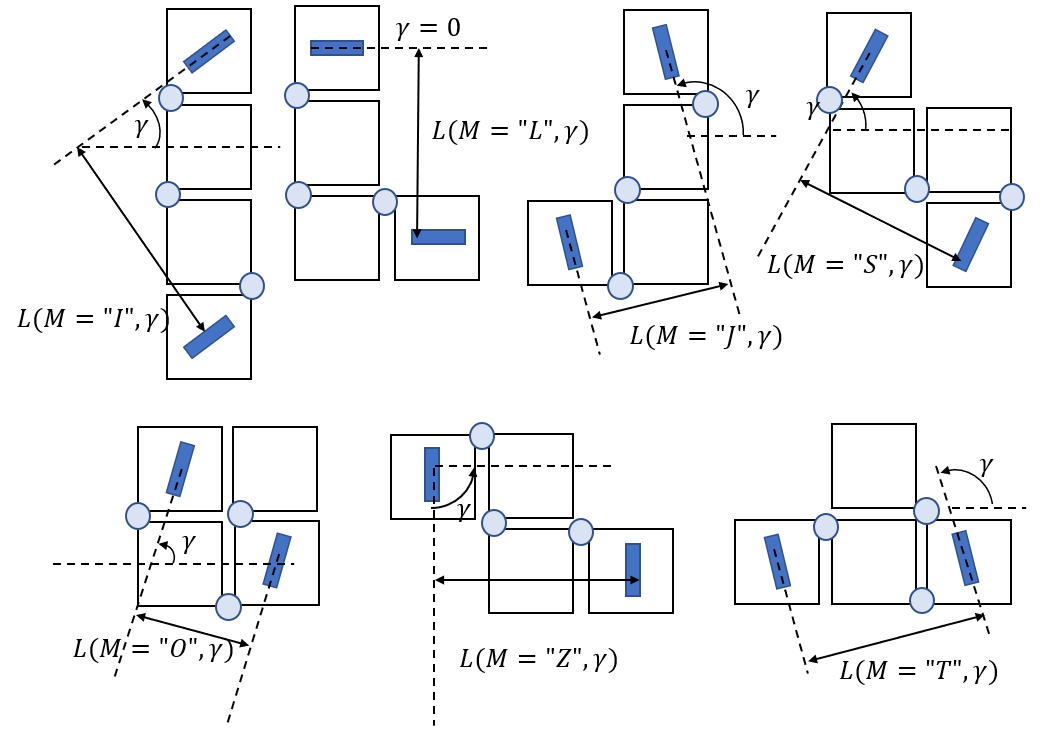}
\caption{Definition of $L(M,\gamma)$ under seven different morphology.}
\label{7examples}
\end{figure}

The position error information is used to generate the angular coordinate of the desired ICR location ($\gamma_d$) which is defined to be
\begin{equation}
    \gamma_d = \arctan{(\frac{Y_e^*}{X_e^*})}    
\end{equation}
It indicates that when there is only an error in the position, $\gamma_d$ will be the driving direction of the robot. On the other hand, when there also exists error in the orientation, the desired radius of rotation is estimated by treating the whole platform as a two wheel differential drive robot. The two wheels with the maximum distance between their x-y planes are modeled to be the two wheels of the differential-drive as shown in Fig. \ref{7examples}. It is because the maximum and the minimum velocity will only occur at these two modules. Thus, taking the two wheels with the maximum distance ensure the desired radius of the curvature does not exceed the hardware limit. Hence, the distance between the two modules is defined to be the width between the two equivalent wheels. Knowing that $\theta_e = \theta_d^*-\theta^*$, the desired angular velocity of the equivalent differential-drive model is controlled by a proportion controller with an adaptive gain $\dot{\theta}_d = k_p\theta_e$. 

The equivalent left wheel velocity ($v'_l$) and right wheel velocity($v'_r$) for the differential-drive model are estimated by
\begin{equation}\label{eqV}
\begin{cases}
$$v'_r = sign(\dot{X}_d^*)\sqrt{\dot{Y_d^*}^2 + \dot{X_d^*}^2} -L(M,\gamma)\dot{\theta}_d$$\\
$$v'_l = sign(\dot{X}_d^*)\sqrt{\dot{Y_d^*}^2 + \dot{X_d^*}^2} +L(M,\gamma)\dot{\theta}_d$$
\end{cases}
\end{equation}
where $L(M,\gamma)$ refers to the perpendicular distance between the equivalent wheels for the particular morphology $M$ with the current driving angle. Also, $sign(\dot{X}_d^*)$ is the signum fuction which gives -1 when $\dot{X}_d^*<0$ or 1 when $\dot{X}_d^*\geq0$. It is used to cover the two dimensional ICR space as $\gamma_d \in [-\frac{\pi}{2},\frac{\pi}{2}]$. Therefore, the instantaneous radius of rotation derived from (\ref{eqV}) can be represented as 
\begin{equation}\label{radius}
R_d  = \frac{L(M,\gamma)}{2}\frac{v'_l+v'_r}{v'_l-v'_r} = \frac{sign(\dot{X}_d^*)\sqrt{\dot{Y_d^*}^2 + \dot{X_d^*}^2}}{\dot{\theta}_d}
\end{equation}
However, there exist a singularity point when $\dot{\theta}_d = 0$ in (\ref{radius}). An alternative method to constrain the radius is proposed below using hyperbolic tangent function with a large value $R_{max}$ considering the resolution of the absolute encoders.
\begin{equation}\label{alterradius}
R_d = R_{max}\tanh(\frac{\sqrt{\dot{Y_d^*}^2 + \dot{X_d^*}^2}}{\dot{\theta}_dR_{max}})
\end{equation}

The location of $ICR_d$ under $\mathcal{F}_b$ is defined by the desired tangential driving angle $\gamma_d$ and the radius $R_d$ which separates the linear motion and the angular motion of hTetro. It allows hTetro platform to correct either of the attributes, i.e., position and orientation while maintaining the other. 

\subsection{Second Layer: Individual Steering Velocity Controller}\label{betaCalculation}

The hTetro is able to change to seven different shapes and the relative position of each locomotion module also changes. The position of the steering axes with respect to $\mathcal{F}_b$ are denoted as ($x^c_i$, $y^c_i$) with $i\in\{1,2,3,4\}$. After that, the desired steering angle ($\beta_{i,d}$) and individual instantaneous radius of rotation ($r_i$) of each module are given by:  
\begin{equation}\label{betaCal}
\beta_{i,d} = \arctan{(\frac{y^c_i-R_d\sin(\gamma_d)}{R_d\cos(\gamma_d)-x^c_i})}-\alpha_j
\end{equation}

\begin{equation}\label{rcal}
r_i = \frac{y^c_i-R_d\sin(\gamma_d)}{\sin(\beta_{i,d})}
\end{equation}
while $\alpha_j = 
\begin{cases}
\alpha_1 \quad i=1\\
0 \quad i = 2\\
\alpha_2 \quad i=3\\
\alpha_2 + \alpha_3 \quad i=4
\end{cases}$

Because the control signals ($\dot{\beta}_{i,d}$) are error driven, when discontinuous signal is given, for instance, when the robot arrived at the waypoint, it is likely that the required motor speed exceeds its hardware limit. Hence, a constrained feedback controller is used to regulate the linear velocities ($v_i$) which is illustrated in Section \ref{Vicalculation}. From the absolute encoder, as shown in Fig. \ref{fig:sublocomotion}, the current steering angle of each locomotion module ($\beta_i$) is used to compare the difference. The initial steering angles are determined according to the $ICR_d$ at that instant. The desired steering velocities are controlled according to the steering error $\beta_{i,e} = \beta_{i,d}-\beta_i$ such that the kinematic constraint are fulfilled during the transition period. In other words, the controller is used to ensure the lines, which are perpendicular to each wheel plane and passes through the centroid of each module, as expressed in (\ref{lines}) are always concurrent when $v_i \neq 0$ because the four locomotion modules are indifferent when $v_i = 0$. The $\alpha_j$ used in (\ref{lines}) are defined the same way as (\ref{betaCal}).
\begin{equation}\label{lines}
-\frac{1}{\tan{(\beta_i+\alpha_j)}}(x-x^c_i)=(y-y^c_i)
\end{equation}
(\ref{lines}) can be written in the form $\mathbf{A}\boldsymbol{x}=\boldsymbol{b}$ where $\boldsymbol{x} = [x \quad y]^T$  and $\mathbf{A}$ is the coefficient matrix of the individual radius of rotation. Then, its augmented matrix 
\begin{equation}
[\boldsymbol{A}|\boldsymbol{b}] =\left [  
\begin{tabular}{cc|c}
 $\frac{-1}{\tan{(\beta_1+\alpha_1)}}$ & $-1$ & $\frac{-1}{\tan{(\beta_1+\alpha_1)}}x^c_1+ y^c_1$  \\
 $\frac{-1}{\tan{(\beta_2)}}$ & $-1$ & $\frac{-1}{\tan{(\beta_2)}}x^c_2+ y^c_2$  \\
 $\frac{-1}{\tan{(\beta_3+\alpha_2)}}$ & $-1$ & $\frac{-1}{\tan{(\beta_3+\alpha_2)}}x^c_3+ y^c_3$  \\
 $\frac{-1}{\tan{(\beta_4+\alpha_2+\alpha_3)}}$ & $-1$ & $\frac{-1}{\tan{(\beta_4+\alpha_3+\alpha_3)}}x^c_4+ y^c_4$  
\end{tabular}
 \right ]
\end{equation}
 With the initialization of the steering angles, it is known that $rank(\mathbf{A}_{init}|\boldsymbol{b}_{init}) = rank(\mathbf{A}_{init}) \leq 2$ which means all four lines that are co-linear with radii are either parallel or concurrent. Similarly, we also know that at the final state all the steering angles should fulfill the kinematic constraint which indicates that $rank(\mathbf{A}_{d}|\boldsymbol{b}_{d}) = rank(\mathbf{A}_d) \leq 2$. Hence, these two situations will be illustrated separately in Section \ref{parallel} and Section \ref{concurrent}.  
 
 \subsubsection{Both initial state and desired state are parallel}\label{parallel}
 Under this situation, $\dot{\theta}_d = 0$ and $\beta_{i,e} = \gamma_d - \gamma$ which means the robot only perform translational motion. The ranks of the coefficient and augmented matrices are one. Therefore, the steering velocity profiles of all four modules should be the same. In this case, a PID controller was used to keep the heading angle unchanged. Therefore, $\dot{\beta}_i(t) = k_p\gamma_e(t) + k_i\int_{0}^{t} \gamma_e(t) dt + k_d\dfrac{d\gamma_e(t)}{dt}$.
  As the initial state with an arbitrary $\gamma$ could satisfy the constraint, substituting $\beta_i$ with $\beta_i + \int_{0}^{t} \dot{\beta}_i(t) dt$ will not violate the constraint as well.
 
 \subsubsection{Either initial state or desired state is concurrent}\label{concurrent}
 When the initial or final state is concurrent, the solution for the system is unique, indicating ranks of both matrices are two. By doing row-echelon reduction and rearranging the column of the the augmented matrix, matrix $\mathbf{A}$ and vector $\boldsymbol{b}$ can be express as (\ref{rowechelon}).
 \begin{gather}\label{rowechelon}
     \mathbf{A}=
     \left [
     \begin{tabular}{cc}
     1 & $\frac{1}{\tan{(\beta_1+\alpha_1)}}$\\
     0 & $\frac{1}{\tan{(\beta_2)}}-\frac{1}{\tan{(\beta_1+\alpha_1)}}$\\
     0 & $\frac{1}{\tan{(\beta_3+\alpha_2)}}-\frac{1}{\tan{(\beta_1+\alpha_1)}}$\\
     0 & $\frac{1}{\tan{(\beta_4+\alpha_2+\alpha_3)}}-\frac{1}{\tan{(\beta_1+\alpha_1)}}$\\
     \end{tabular}
     \right ]\\
     \boldsymbol{b} = 
     \left [
     \begin{tabular}{c}
     $\frac{1}{\tan{(\beta_1+\alpha_1)}}x^c_1+y^c_1$\\
     $\frac{1}{\tan{(\beta_2)}}x^c_2-\frac{1}{\tan{(\beta_1+\alpha_1)}}x^c_1+y^c_2-y^c_1$\\
     $\frac{1}{\tan{(\beta_3+\alpha_2)}}x^c_3-\frac{1}{\tan{(\beta_1+\alpha_1)}}x^c_1+y^c_3-y^c_1$\\
     $\frac{1}{\tan{(\beta_4+\alpha_2+\alpha_3)}}x^c_4-\frac{1}{\tan{(\beta_1+\alpha_1)}}x^c_1+y^c_4-y^c_1$\\
     \end{tabular}
     \right ]
 \end{gather}
 
 Due to the geometry property of the seven configurations, the relationship among all $x^c_i$ or among $y^c_i$ are summarized in TABLE.\ref{xciycirelationship}.

 \begin{table}[!t]
     \centering
      \vspace*{0.1in}
     \caption{Relationships among $x^c_i$ or $y^c_i$}
     \label{xciycirelationship}
     \begin{tabular}{c|c|c}
     \hline
    shape & $x^c_i$ relationship & $y^c_i$ relationship \\
    \hline \hline
     I & $x^c_1=x^c_2=x^c_3=x^c_4=0$ & $y^c_1=3y^c_2=-3y^c_3=-y^c_4$ \\
     L & $x^c_1=x^c_2=x^c_3=-x^c_4$ & $y^c_1=3y^c_2=-3y^c_3=y^c_4$ \\
     Z & $x^c_1=x^c_4,x^c_2=x^c_3=0$ & $y^c_1=y^c_2=-y^c_3=-y^c_4$ \\
     O & $x^c_1=x^c_2=-x^c_3=-x^c_4$ & $y^c_1=-y^c_2=-y^c_3=y^c_4$ \\
     T & $x^c_1=-3x^c_2=x^c_3=x^c_4$ & $y^c_1=-y^c_4,y^c_2=y^c_3=0$ \\
     S & $x^c_1=x^c_3=0,x^c_2=-x^c_4$ & $y^c_1=y^c_2=-y^c_3=-y^c_4$ \\
     J & $-x^c_1=x^c_2=x^c_3=x^c_4$ & $y^c_1=y^c_2=-3y^c_3=-y^c_4$ \\
     \end{tabular}
     \label{tab:my_label}
 \end{table}
 
 As a result, by substituting the relationships in TABLE.\ref{xciycirelationship} to (\ref{rowechelon}), it can be found that the following relationship between the steering angles should be fulfilled. 
 \begin{multline}\label{steeringRelationship}
     \frac{1}{\tan{(\beta_1+\alpha_1)}}-C_1\frac{1}{\tan{(\beta_2)}}-C_2\frac{1}{\tan{(\beta_3+\alpha_2)}}\\
     -C_4\frac{1}{\tan{(\beta_4+\alpha_2+\alpha_3)}}=0
 \end{multline}
 The relationship above is derived by assuming the system is consistent. $C_i$ are constant coefficient for different shapes with respect to TABLE.\ref{xciycirelationship}. Due to the assumption made in Section \ref{kinematics}, the relationships among $x^c_i$ and $y^c_i$ will not change during locomotion. Thus, in order to fulfill the concurrent requirement, $\dot{\beta}_i(t)$ should be restricted by the following constraints as expressed in \ref{c1} and \ref{c2}. 
 \begin{equation}\label{c1}
 \int_{0}^{t_{final}} \dot{\beta}_i(t) dt = \beta_{i,e} 
 \end{equation}
 This constraint is to ensure it reaches the desired angle at the same time instance $t_{final}$.
 \begin{equation}\label{c2}
 \tan{(\beta_{i,init}+\int_{0}^{t} \dot{\beta}_i(t) dt+\alpha_j)}=\lambda\tan{(\beta_{i,init}+\alpha_j)}
 \end{equation}
 or 
 \begin{equation}\label{c3}
 \tan{(\beta_{i,init}+\int_{0}^{t} \dot{\beta}_i(t) dt+\alpha_j)}=\lambda\tan{(\beta_{i,d}+\alpha_j)}
 \end{equation}
 This is to ensure the kinematic constraint during the transient state depending on the concurrent relationship. 
 
 Initially, $\dot{\beta}_i^+$ is defined to be the desired velocity of the first time instance.
 \begin{equation}\label{betaplus}
 \beta_i^+ = \int_{0}^{t^+} \dot{\beta}_i^+ dt
 \end{equation}
 Substitute (\ref{betaplus}) into (\ref{c2}) and $t=t^+$,
 \begin{gather}
      \beta_{i,e}^+ = \beta_{i,e}-\beta_i^+ \\
     \tan{(\beta_{i,init}+\beta_i^++\alpha_j)}= \frac{\tan{(\beta_{i,init}+\alpha_j)}+\tan{(\beta_i^+)}}{1-\tan{(\beta_{i,init}+\alpha_j)}\tan{(\beta_i^+)}}
 \end{gather}
 Hence,
 \begin{equation}\label{lambda}
     \lambda = \frac{1+\frac{\tan{(\beta_i^+)}}{\tan{(\beta_{i,init}+\alpha_j)}}}{1-\tan{(\beta_i^+)}}= \frac{1+\frac{\tan{(\beta_i^+)}}{\tan{(\beta_{i,d}+\alpha_j)}}}{1-\tan{(\beta_i^+)}}
 \end{equation}
In order to ensure the linear dependency is followed by all four module, $\lambda$ is defined by substituting the maximum $\dot{\beta}_i$ in to (\ref{betaplus}) which is determined by the maximum $\beta_{i,e}$ to fulfill criteria (\ref{c1}) through a PID controller. Thus, the module with the maximum error will be used as reference to calculate the $\lambda$. After that, the rest of the steering angles ($\beta_i^+$) are calculated based on (\ref{lambda}) which is then used to determine the steering speed of the rest modules ($\dot{\beta}_i$).
 
\subsection{Constraint Individual Velocity Controller}\label{Vicalculation}
According to Section \ref{control}, a constraint velocity controller need to be implemented to restrict the maximum drive rate of each motor. Referring to (\ref{rcal}) and (\ref{radius}), the desired linear velocity of each module can be expressed as
 
 \begin{equation}\label{vicalculation}
     v_{id} = \frac{r_i}{R_d}\sqrt{\dot{Y_d^*}^2+\dot{X_d^*}^2} = \frac{r_i}{R_d}\dot{\theta}_d=\frac{r_i}{R_d}k_p(\theta_d-\theta)
 \end{equation}
 
the desired drive rate of each wheel motor for each module i is given as:
 \begin{equation}\label{driverate}
     \dot{\phi}_{iL} = \frac{\dot{\beta_i}}{r_w}+\frac{\dot{v}_{id}}{r_w}, \quad \dot{\phi}_{iR} = \frac{\dot{\beta_i}}{r_w}-\frac{\dot{v}_{id}}{r_w}
 \end{equation}
 Substituting (\ref{vicalculation}) into (\ref{driverate}),
 \begin{gather}\label{driverate2}
     \dot{\phi}_{iL} = \frac{\dot{\beta}_id}{r_w}+\frac{\frac{r_i}{R_d}k_p(\theta_d-\theta)}{r_w}\\
     \dot{\phi}_{iR} = \frac{\dot{\beta}_id}{r_w}-\frac{\frac{r_i}{R_d}k_p(\theta_d-\theta)}{r_w}
 \end{gather}
 
 In order for the drive rate not to exceed the drive limit $\dot{\phi}_{max}$ according to the hardware limitation of the motor, $k_p$ is determined by limiting the larger velocity between (19) and (20) to $\dot{\phi}_{max}$.
 
 \begin{equation}
     k_p = \frac{r_w\min{\big\{\dot{\phi}_{max},\max{\{\dot{\phi}_{iL},\dot{\phi}_{iR}\}}\big\}}- \dot{\beta}_i}{\frac{r_i}{R_d}(\theta_d-\theta)}
 \end{equation}
 where the $\max$ function is used to extract the maximum desired drive rate among all eight motors and the $\min$ function is used to compare it with the limit. After determining the gain value, the real driving command $\dot{\phi}_{i,L}^{real}$ and $\dot{\phi}_{i,R}^{real}$ can be obtained by substituting the $k_p$ into (\ref{driverate2}) which will be fed into the platform.
\section{EXPERIMENTS AND RESULTS}\label{experiment}
\begin{figure}
    \centering
     \vspace*{0.1in}
    \includegraphics[width=3.2in]{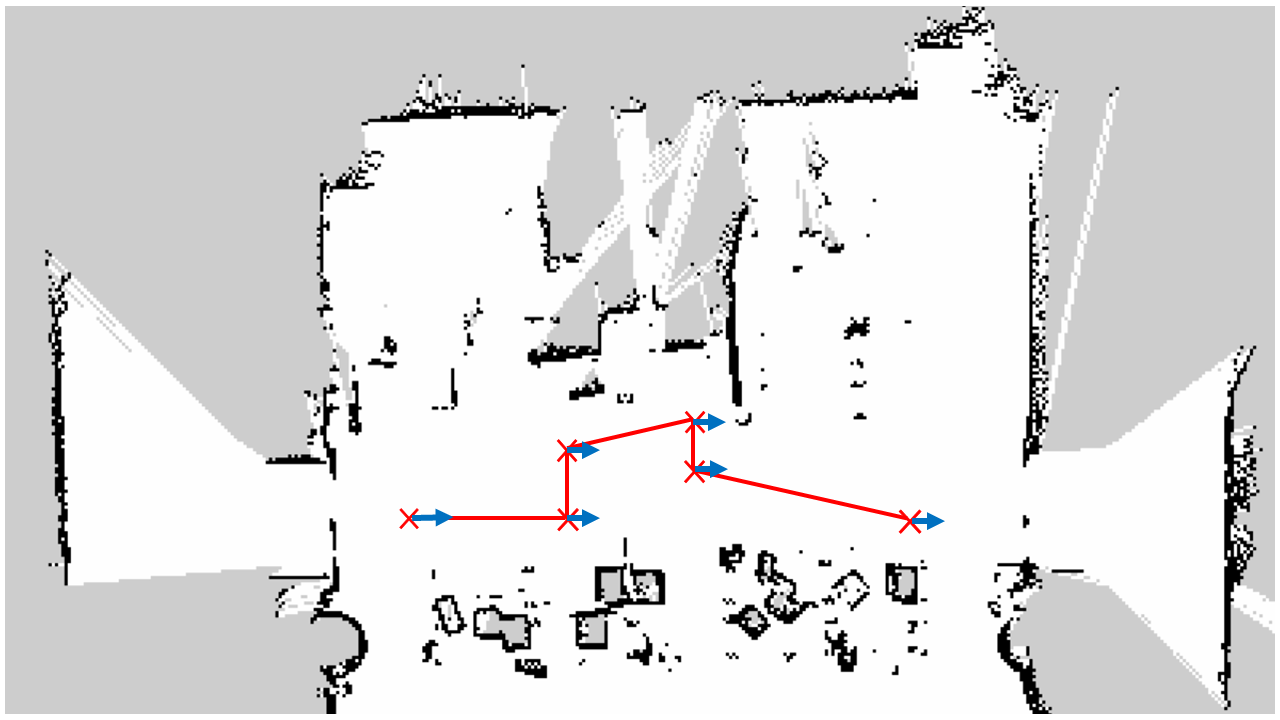}
    \caption{2D map used for the experiment}
    \label{fig:map}
\end{figure}
In this section, the experimental setup with explanation is presented in detail, together with the results and discussions. The experiments were done using the hTetro platform as shown in Fig. \ref{fig_hardware}. Fig. \ref{fig:map} shows the 2D map used for the experiment. The waypoints set for the robot are marked by red crosses. 
The same path is tested using all seven configurations and the results are discussed.

\begin{figure*}[!t]
\centering
\includegraphics[width = 6.9in]{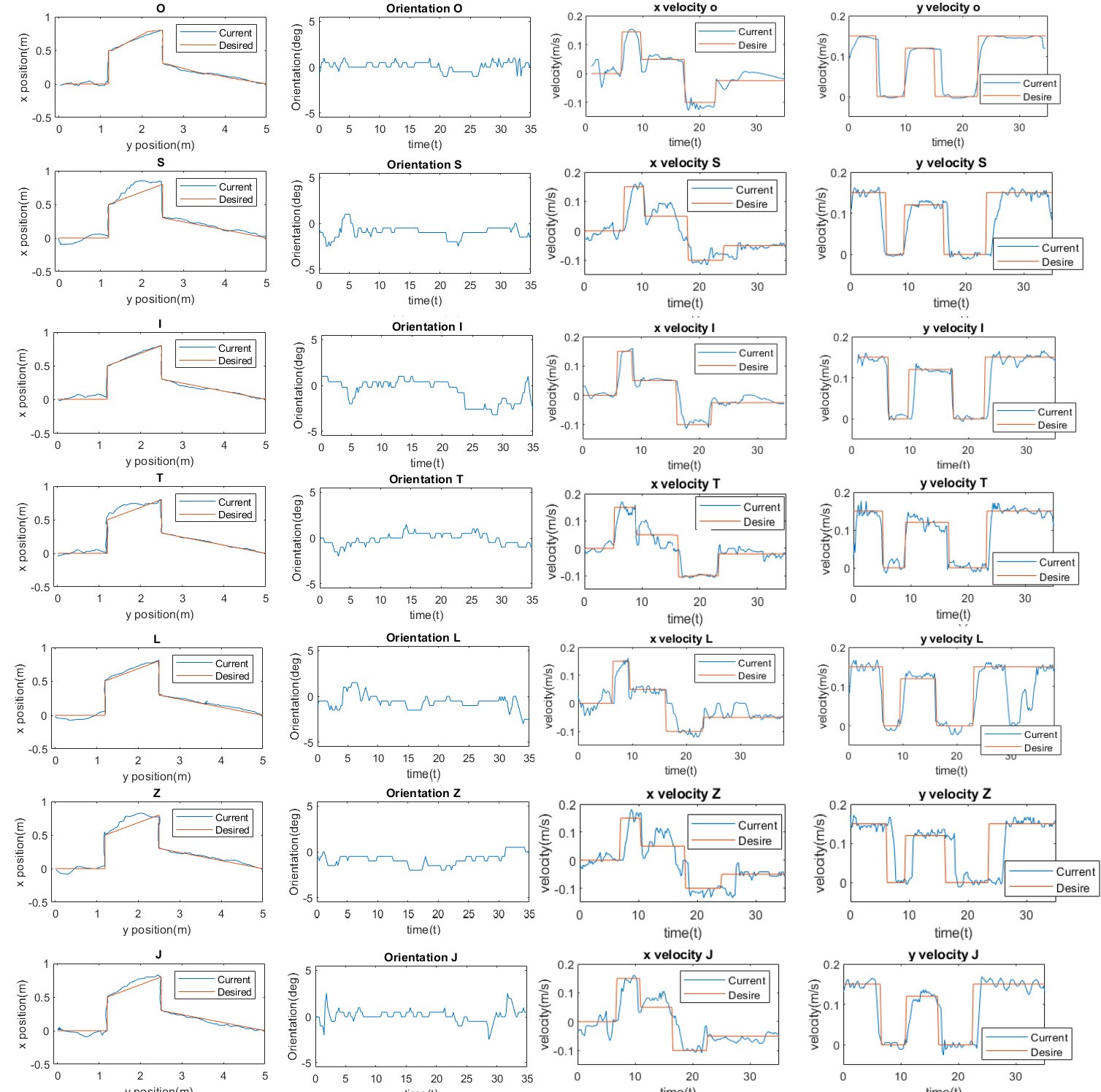}
\caption{Results of the proposed controller for each configuration of hTetro under Cartesian space.}
\label{fig_positionresult}
\end{figure*}

In order to test it in the real scenario, only waypoints containing the desired position and orientation are given to the robot. The waypoints are set to make the robot perform the zig-zag pattern and diagonal motion while keeping the orientation and steering angles with the controller proposed above. Sequences of the waypoints received by the robot are (0, 1.2), (0.5, 1.2), (0.8, 2.5), (0.3, 2.5), and (0, 5) with all dimensions measured in meters. 

As shown in the Fig. \ref{fig_positionresult}, all seven configurations of the platform demonstrated the ability to follow the desired trajectory based on discrete waypoints and velocity commands under the Cartesian space. From the data collected, the minimum root mean square errors of the positions in $X$ and $Y$ are implicitly 0.108m and 0.0623m which is from 'O' configuration, while the maximum root mean square errors are 0.129m and 0.16m which from 'S' configuration. The suspected reason for the weak performances in 'S', 'T', and 'Z' shapes is the non-linearity of the absolute encoder. Due to the phase difference of $\frac{\pi}{2}$ between two modules, the desired angles fall into the non-linear region of the absolute encoder and hence causing deviations. The ability to recover from disturbances is proved with the supplementary video.\footnote{ https://1drv.ms/v/s!AuXDySZjV7Lddb44kt2bOHUA0Fg?e=1Efijq}
In addition, when implementing the control strategy on the platform, some overshoot and oscillation response can be observed in the velocity plot. The first reason we suspected is the joggle of the robot chassis which resulted in oscillation of the LIDAR. Secondly, imperfect tuning parameters of the motor controller might be another reason that would result in the oscillation in the velocity plot. These two issues are technical problems related to the fabrication of the platform.
\section{CONCLUSIONS}\label{conclusion}

In this paper, we presented the kinematic modeling and path tracking controller design of a novel modular reconfigurable cleaning robot with the usage of four differential drive modules. The usage of differential drive units to replace individual steering drive units helps to avoid the kinematic singularity. Kinematics modeling was done considering the self-reconfiguration of the hTetro robot into seven forms. The mapping of the ICR location enables the robot to decouple the linear and angular motion under Cartesian space. Additionally, the velocity controllers regulate the kinematic and hardware constraints of the platform making it able to fulfil the kinematic constraints under the transition period. Experiments on path tracking and overcoming disturbances were performed. The results reflect the ability of trajectory following and fast recovery from disturbances. Future research will focus on improvement on the robustness of platform, and dynamic controller design considering the influence of the cleaning brushes.

\section{ACKNOWLEDGMENT}

Anonymous reviewers of the manuscript are greatly acknowledged for their valuable comments which improved the manuscript significantly. We are thankful to Dr. Abdullah Aamir Hayat at Singapore University of Technology and Design(SUTD), Singapore for proofreading the manuscript. This project is supported by Singapore National Robotics R\&D Program Office under the Grant NO. RGAST1702 and SUTD-MIT International Design Center.
\bibliographystyle{IEEEtran}
\bibliography{reference}

\end{document}